\documentclass{article}

\usepackage[nonatbib, final]{neurips_2020}

\usepackage[utf8]{inputenc} % allow utf-8 input
\usepackage[T1]{fontenc}    % use 8-bit T1 fonts
\usepackage{hyperref}       % hyperlinks
\usepackage{url}            % simple URL typesetting
\usepackage{booktabs}       % professional-quality tables
\usepackage{amsfonts}       % blackboard math symbols
\usepackage{nicefrac}       % compact symbols for 1/2, etc.
\usepackage{microtype}      % microtypography
\usepackage{graphicx}
\usepackage{todonotes}
\usepackage{tikz}           % labeling X and Y axis
\usetikzlibrary{positioning}
\usepackage{multirow}

\usepackage{cleveref}
\usepackage[square,sort,comma,numbers]{natbib}

\title{Hybrid Backpropagation Parallel Reservoir Networks}

\author{%
    Matthew S. Evanusa \thanks{Corresponding author. Code available at: https://github.com/Symbiomancer/HBP-ESN} \\
    Department of Computer Science\\
    University of Maryland\\
    College Park, MD \\
    \texttt{evanusa@cs.umd.edu}
    \And
    Snehesh Shrestha \\
    Department of Computer Science\\
    University of Maryland\\
    College Park, MD \\
    \And
    Michelle Girvan \\
    Department of Physics\\
    University of Maryland\\
    College Park, MD
    \And
    Cornelia Ferm\"uller\\
    Institute for Adv. Computer Studies\\
    University of Maryland\\
    College Park, MD\\
    \And
    Yiannis Aloimonos\\
    Department of Computer Science\\
    University of Maryland\\
    College Park, MD\\

}

\begin{document}

\maketitle

\begin{abstract}
  %Learning data that evolves in time is still an open area of research, with recurrent neural networks, neural networks that contain cycles which are inspired by the massive amount of recurrence in the brain, being a promising tool.  

  In many real-world applications, fully-differentiable RNNs such as LSTMs and GRUs have been widely deployed to solve time series learning tasks.  These networks train via Backpropagation Through Time, which can work well in practice but involves a biologically unrealistic unrolling of the network in time for gradient updates, are computationally expensive, and can be hard to tune.  A second paradigm, Reservoir Computing, keeps the recurrent weight matrix fixed and random.  Here, we propose a novel hybrid network, which we call Hybrid Backpropagation Parallel Echo State Network (HBP-ESN) which combines the effectiveness of learning random temporal features of reservoirs with the readout power of a deep neural network with batch normalization.  
  We demonstrate that our new network outperforms LSTMs and GRUs, including multi-layer "deep" versions of these networks, on two complex real-world multi-dimensional time series datasets: gesture recognition using skeleton keypoints from ChaLearn, and the DEAP dataset for emotion recognition from EEG measurements.  We show also that the inclusion of a novel meta-ring structure, which we call HBP-ESN M-Ring, achieves similar  performance  to one large reservoir while decreasing the memory required by an order of magnitude. We thus offer this new hybrid reservoir deep learning paradigm as a new alternative direction for RNN learning of temporal or sequential data.

\end{abstract}

\section{Introduction and Motivations}
\label{S:1}

Temporal learning is still a hard task that currently requires massive network architectures to achieve satisfactory results.  In order to process temporal information with neural networks, the network needs to keep a memory of its past activity.  This memory can be accomplished by either feeding the entire sequence into a feed-forward network, as in the style of transformer-attention architectures \cite{vaswani2017attention}, or by feeding back the activity from the previous state into the neuron at the next time step. The main drawback of Transformer networks is that they require the entire sequence be passed in at once, and this prevents  online learning.  The domains of deep learning and reservoir computing need not be mutually exclusive, and years of research into optimizing deep networks can benefit the reservoir paradigm.
%and means that transforms are not a class of RNN.
%Thus they are not as well suited as RNNs for online or real-world time series learning.  
Here, we focus on a novel RNN architecture and compare it to other state-of-the-art RNN architectures that take in input one item at a time, rather than as the entire sequence. 

\subsection{Engineering Motivation}

A fundamental assumption in all learning tasks is that the 'real' world from which we  sample is a dynamical system -- each state of the latent variables of the outside world can be described by a differential equation given the last state. The feed-forward approach to learning temporal sequences amounts to a sampling of the sequence at regular intervals, taking advantage of Takens' theorem \cite{jaeger2001echo}, which  says that a dynamical system can be approximated using a sufficient number of samples of the system in a feed-forward network.  If one chooses to learn temporal sequences with a feed forward network in this manner, however,  the model size must be drastically increased to account for each temporal step, and arbitrary-length sequences cannot be used.  The feed forward network must act as an unrolled recurrent network. 

%Any feed-forward neural network is a function: it maps its inputs in a non-linear fashion to an output.  By contrast, a recurrent neural network of any form is itself a \emph{dynamical system} \cite{lukovsevivcius2009reservoir}, which is capable of adapting itself to the dynamics of the input [cite]. 

A recurrent neural network compacts a deep feed-forward network into a more efficient representation \cite{van2020going}. However, the issue then becomes how to learn reusable recurrent weights for an aribtrary sequence.  This work proposes a novel method that avoids gradient-based updates in the recurrent layers.  Here we argue that reservoir-type recurrent networks offer an efficient manner of temporal processing \emph{without} having to unroll the network, and the complications that come with learning on the unrolled network.  

\subsection{Biological Motivation}

In addition to power consumption and metabolic limitations, it has been argued that biological brains also 'compact' the unrolled network into a recurrent one \cite{van2020going}. Beyond just merely having recurrent connections, there is a biological argument that human brain regions actually behave as a reservoir \cite{enel2016reservoir}, and that reservoir computing can offer a method to study cortical dynamics \cite{singer2013cortical}.  Furthermore, \cite{russindeep} argued that deep learning is lacking  concepts inherent to the  prefrontal cortex, and one of them  is the existence of recurrent connections. 

It is important to note that we do not argue that the current architecture is fully biologically plausible, because it employs backpropagation. However, backpropagation here is used only as a readout mechanism (see section 2), and in the future it could be replaced with any  suitable supervised learning mechanism that can learn non-linear classifications, including any biologically-realistic ones; this does not effect the reservoir mechanism.  

Our main philosophy is  that engineering and biological needs are not mutually exclusive \cite{van2020going} and can inform each other in the generation of new learning algorithms, especially when the end goal of A.I. is a learning system that acts and behaves like human intelligence \cite{lake2017building}.
%We take the philosophy here that trying to build bridges between deep learning and neuroscience can yield benefits for both communities, even if certain parts are not yet fully biologically realistic.  In fact, it is hard or impossible at this point to fully tell what is in fact "biologically realistic".  

\subsection{Recurrent Neural Networks}
\subsubsection{Types of Recurrent Networks}

Allowing the network to learn not just static features, but also temporal information, increases the learning capacity of the network
%allowing the  networks to learn entire sequences rather than through the required frame-based learning that comes with feed-forward architectures
\cite{walter2016computation, izhikevich2004spike}.  Within the recurrent neural network (RNN) umbrella, we have two paradigms: a) fully differentiable recurrent networks, such as LSTMs \cite{hochreiter1997long}, and b) Reservoir Computing (RC).  We will not discuss here another paradigm, Spiking Neural Networks (SNNs) \cite{ghosh2009spiking}, which are by original design not a purely recurrent architecture, but can be made one (see below for LSMs), which muddies the classification. 
%We limit our discussion to reservoir computers using rate encoding, as opposed to spikes as in the LSMs.
%and muddy the classification; the distinction there is between rate-encoding and spiking neurons. The classification here is strictly between fully and non-fully differentiable RNNs.  
In fully differentiable RNNs, the learning mechanism required needs to handle backpropagating errors through the recurrent loops, and so algorithms such as unrolling backpropagation (BPTT) are currently the most widely used \cite{werbos1990backpropagation}, although newer alternatives to BPTT do exist \cite{bellecbiologically}.  In reservoir computing, the recurrent connections are kept random and fixed, and only the readout of the reservoir is trained, i.e., the states of the reservoir are mapped to targets using a learning algorithm ; this removes the need to backpropagate the error through the recurrent layers.  Reservoirs can come in one of two "flavors": rate encoding real-valued reservoirs, Echo State Networks (ESN) \cite{jaeger2001echo}, and spike or event-encoding Liquid State Machines (LSM) \cite{maass2011liquid}. 

\subsubsection{Learning with Random RNNs}

One can view the reservoir as acting as a temporal kernel \cite{tino2020dynamical}, expanding out a temporal sequence into a high dimensional random space that makes it more separable (linearly or otherwise).  This temporal kernel works well even though the weights are kept random, which is similar to theories of the brain that are built off of unsupervised organizations of neuronal groups \cite{edelman1993neural}. Once the reservoir expands out the data, it is traditional to use methods such as ridge or lasso regression to learn to map reservoir states to targets.  This work replaces such a readout with a deep-network backpropagation mechanism.  Reservoir computing has been shown to train faster and use less training data than their fully-differentiable counterparts in some domains \cite{neofotistos2019machine}. For real world tasks, ESNs have been successfully used for a variety of tasks ranging from direct robotic control \cite{polydoros2015advantages}, electric load forecasting \cite{bianchi2015short}, wireless communication \cite{jaeger2004harnessing}, image classification \cite{tong2018reservoir}, robot navigation \cite{antonelo2014learning},  reinforcement learning \cite{chang2019convolutional}, and recently action recognition using preprocessed inputs \cite{soures2019deep}.  %In this work, our task will be a gesture recognition classification using preprocesed video skeleton keypoints. 

In introducing ESNs, Jaeger \cite{jaeger2001echo} wrote that a potential limiting factor of using ridge regression for ESNs is that it is only a linearly separable readout, and that this might be limiting the learning capabilities of the system. This motivates an investigation into whether using readout mechanisms that can learn non-linear mappings can improve learning. In this work, we will show that using a readout capable of learning non-linear mappings in conjunction with a parallel structure greatly improves classification and regression accuracy.  

One way to view reservoir computers is through the physics paradigm, where the network is learning a dynamical attractor state (cite Herbert 2004).  From a neuroscience perspective, however, the reservoir can be seen through the lens of the "assembly readout mechanism" paradigm \cite{buzsaki2010neural}, where the reservoir is producing dynamical features, and the readout mechanism is interpreting them. 

\subsection{Contributions}
This work is the first work, to the best of the authors' knowledge, that combines a novel shared-weight parallel ring reservoir scheme, a backpropagated readout mechanism, and modern normalization techniques, and compares it with a state of the art fully differentiable stacked LSTM and GRU, on two  real-world learning tasks.  It is  also the first work to perform rigorous testing on the efficacy of the backpropagated mechanism versus traditional ridge regression mechanisms.  Our hope is to bring together advances in reservoir computing from the physics community, as well as advances from the deep learning community in machine learning, and combine them to greater effect. 

Our contributions to the current state of the art are: %a combination of: 
\begin{itemize}
\item Use of a parallel reservoir mechanism in conjunction with a backpropagated deep feedforward network readout, what we call a Hybrid Backpropagation Parallel Echo State Network (HBP-ESN) and a variant with a shared interconnection weight matrix (HBP-ESN M-Ring)
\item Testing of the HBP-ESN and HBP-ESN M-Ring on two challenging and different real-world datasets, a gesture recognition classification task and EEG signal emotion regression task
\item Analysis and use of modern normalization techniques such as weight decay and batch normalization
\item Experimentally demonstrate that the network can significantly outperform an LSTM and GRU 
\end{itemize}

%\begin{enumerate}
%\item Numbered list item one
%\item Numbered list item two
%\end{enumerate}

\begin{figure}[htbp]
\begin{center}
\includegraphics[width=\textwidth,height=\textheight,keepaspectratio]{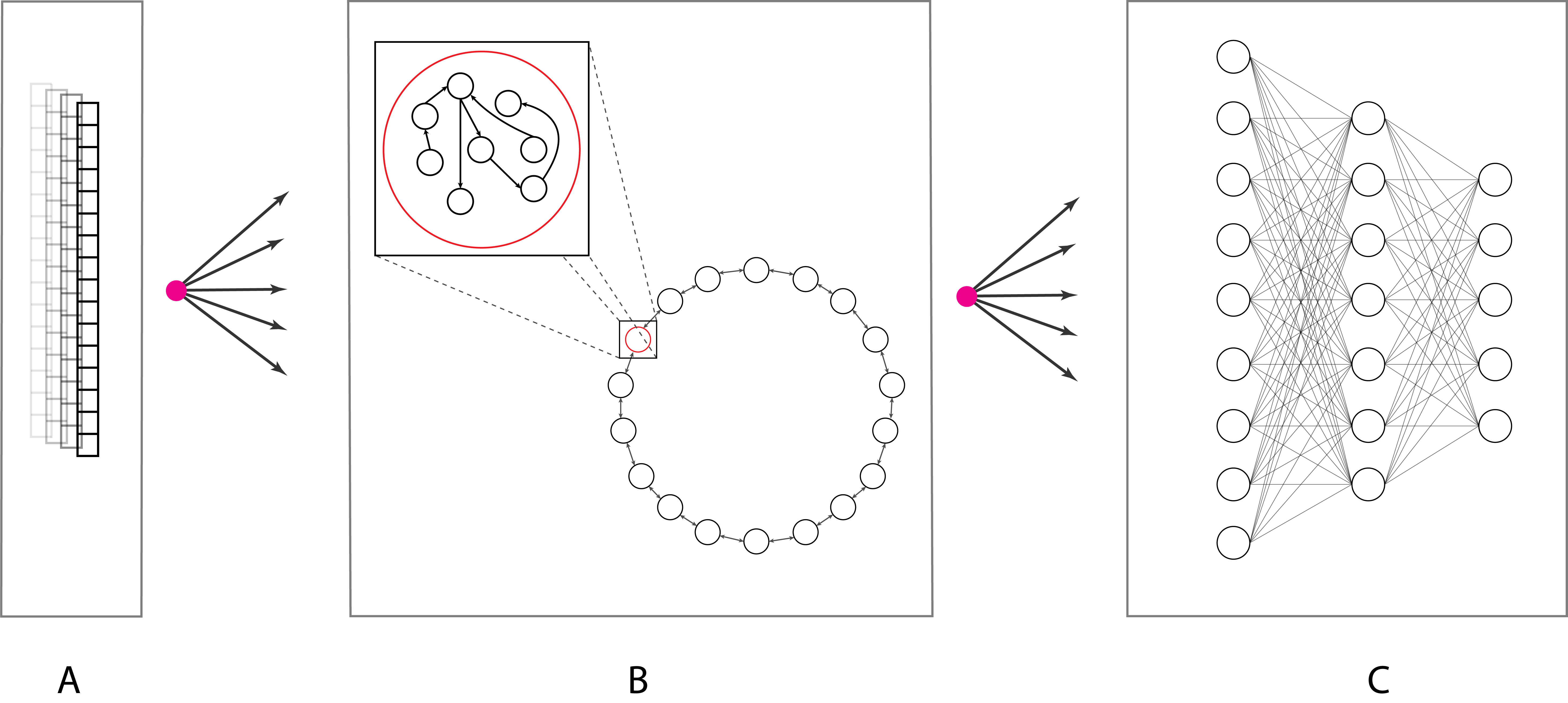}

\caption{The HBP-ESN M-Ring architecture:  (A) the input to the reservoir, with shading corresponding to time steps;  (B) the parallel ring reservoir;  (C) the backpropagated readout. The magenta dots and fan-out arrows between A, B, and C represent fully connected matrices.  The ring in B is straightened and then flattened before being passed into C.  Each sub-reservoir, shown zoomed in, is connected to its left and right neighbor in the shared weights structure, forming a ring of sub-reservoirs.  In different experiments, some elements are swapped; shown here is the full contribution with the shared parallel cross-talk weights.  HBP-ESN without M-Ring lacks the ring links. In the single reservoir experiment, part B consists solely of a large version of the zoomed-in sub reservoir.  For the feed-forward only experiment, part B is removed entirely.}
\end{center}
\label{fig:M-ring}
\vspace{-8mm}
\end{figure}

%\section{BPR-ESN Architecture}
\section{HBP-ESN Architecture}

\subsection{Reservoir Architecture}

The most standard way to implement a reservoir computer is to attach a set (vector) of recurrent, randomly weighted neurons that remain fixed and unlearned, to a readout mechanism such as ridge regression, or sometimes, stochastic gradient descent [cite].  Here, we use the nonlinear leaky updates from \cite{tong2018reservoir}.  At each timestep $t$, input is fed into the reservoir through a weight matrix $W_{in}$, and the reservoir and output vectors are updated according to the following equations:

\begin{equation}
  %  \begin{centered}
    x_t = (1-\alpha)x_{t-1} + \alpha f(u_tW_{in} +  x_{t-1}W_{rec})
   % \end{centered}
\end{equation}

\begin{equation}
  %  \begin{centered}
    y_t = x_t W_{out},
 %   \end{centered}
\end{equation}

where $y_t$ is the output vector at time $t$, $x_t$ is the $N$-dimensional vector of the states of each reservoir neuron at time $t$, $u_t$ is the input vector at time $t$, $\alpha$ is the leak rate, $f(\cdot)$ is a saturating non-linear activation function (here hyperbolic tangent), and $W_{in}, W_{rec}, W_{out}$ are the input-to-reservoir, reservoir-to-reservoir, and reservoir-to-output weight matrices, respectively.  If the input vector has length $A$, and the output vector has length $B$, then the dimensions of $W_{in}, W_{rec}, W_{out}$ are $N\times A$, $N \times N$, $B \times N$ respectively. The ordering of the weight matrix multiplication is reversed based on whether one is using row-major or column-major weight ordering. From the updates, one can see that the reservoir keeps a memory trace of its past activity through the recurrent weights $W_{in}$, which is a function of the leak rate $\alpha$ as well as the spectral radius of $W_{in}$ [cite]. In order to make sure inputs vanish over time, it is necessary to ensure the \emph{echo state property} of the network, which says that the spectral radius, or largest singular value of the recurrent matrix, is less than one.  Although this property has been re-analyzed \cite{yildiz2012re}, for simplicity we will use the original recipe for calculating the desired spectral radius as in \cite{jaeger2001echo}.  For simplicity of the model and ease of replication, we also do not hyper-tune the reservoir weights using any optimization techniques such as genetic algorithms \cite{ferreira2009genetic, ferreira2011comparing} or particle swarm optimization \cite{chouikhi2017pso, basterrech2014experimental}, although those can readily be applied to this work to decrease the variance of the reservoir trials.

The paradigm of reservoir computing simply prescribes for three elements: a read-in of the input, a reservoir, and a read-out that learns the states of the reservoir.  It is the hope, but not guaranteed, that the reservoir will act as a perfect temporal kernel, expanding the input into a higher dimensional space that is \emph{linearly} separable by the readout mechanism.  However, often this may in fact not be the case for all possible inputs
%, for a given reservoir.
One way to show this is to compare the performance of a linear learning method versus a non-linear method, such as a deep neural network, which is what we attempt to do here.  We want to tease apart the three components into their own modules; for the purposes of this current work, we swap out the readout module. 

\subsection{Parallel Reservoir Structure}

\subsubsection{Motivations}

As in \cite{pathak2018model}, we implement a parallel reservoir structure.  This structure serves a number of purposes.  Biologically, the brain is partitioned into multiple sub-regions, which aids in its functions, for example see \cite{rypma1999roles, jensen2007separate}.  On the other hand, there exist networks with massive cross-connectionism such as in Hopfield networks \cite{mackay2003information}, which have the issue of what is known as "catastrophic interference" \cite{french1999catastrophic}, wherein learning new information causes old information to be lost.  Of course, we are not learning the weights of the recurrent network and thus cross-contamination is less of an issue, %versus Hopfield networks,
however we take this simply as motivation.  By partitioning the networks, we prevent information from over-saturating other regions and enforce some sub-structural components.  This catastrophic forgetting occurs not just in Hopfield networks, but in feed-forward MLP networks as well. There is empirical evidence that partitioning a network can alleviate this catastrophic forgetting \cite{masse2018alleviating}.  
%For reservoir computers, 
A biologically-inspired reservoir network was shown to improve its capabilities by partitioning activity using gating \cite{rikhye2018thalamic}. 

Second, by separating into smaller sub-reservoirs, our hope is that we can decrease the exponential increase of the size of one large reservoir.  As our experiments show and as in \cite{pathak2018model}, multiple small reservoirs perform well in the face of an increase of the size of the input.  Third, multiple reservoirs are better adept at handling data that is naturally already partitioned.  In our experimental data, the vector is already partitioned into (x,y) components, with each two indexes comprising one (x,y) position.  This concept can be extended to any case where the data naturally partitions itself, such as regions of an image.  Due to the shared connectivity matrix described in section 2.2.3, the sub-reservoirs are able to access information about other regions as well. 

\subsubsection{Implementation}

Equation 1 becomes, for the parallelized scheme without the cross-talk matrix:

\begin{equation}
   % \begin{centered}
    x_{t,r} = (1-\alpha)x_{t-1,r} + \alpha f(u_tW_{in,r} +  x_{t-1,r}W_{rec,r})
 %   \end{centered}
\end{equation}

and for the shared cross talk matrix:

\begin{equation}
  %  \begin{centered}
    \delta_{ring} =  \beta (f(x_{t-1,r-1}W_{shared}) + f(x_{t-1,r+1}W_{shared}))
  %  \end{centered}
\end{equation}

\begin{equation}
   % \begin{centered}
    x_{t,r} = (1-\alpha)x_{t-1,r} + \alpha f(u_tW_{in,r} +  x_{t-1,r}W_{rec,r}) + \delta_{ring}
  %  \end{centered}
\end{equation}

Equation 2 becomes, for both cases: 

\begin{equation}
%    \begin{centered}
    y_t = [x_{t,1};x_{t,2};...;x_{t,R}] W_{out}
 %   \end{centered}
\end{equation}

for all sub-reservoirs $r$ from $1$ to $R$, where $R$ is the number of sub-reservoirs, and ';' is the concatenation operator in eq. 4.  All sub-reservoirs are stored in an $R$ dimensional array, referred to henceforth as the sub-reservoir array. For equation 4, reservoir indexes -1 and +1 reflect the left and right index in the sub-reservoir array. The sub-reservoir connections are wrapped around, forming a ring structure.  $\beta$ is a parameter chosen to reflect how strongly to weight the cross-talk connections.  $W_{shared}$ is the shared connectivity matrix used for all sub-reservoirs, which saves on memory usage. We use the same alpha for each sub-reservoir, although this can be parameterized as well.  
%For simplicity, we do not consider cross-talk between sub-reservoirs as in \cite{pathak2018model}, although this can be a topic for future work. 

A significant benefit of the parallel reservoir structure is a dramatic decrease in the memory usage of the weight matrix, which alleviates the $N^2$ increase in the size of the weight matrix for the recurrent weights.  As an example, for a large single-reservoir of 3200 neurons, we would have $3200^2$ recurrent weights, on the order of $10^7$.  However, for he same number of total neurons organized into 8 sub-reservoirs with 400 neurons each, this only amounts to $4 \times 400^2$, or on the order of $10^6$, a full order of magnitude smaller. This would allow for large scaling of network sizes.  The use of a shared cross-talk matrix, $W_{shared}$, further cuts down on memory usage: this matrix needs to be allocated only once, and has  constant memory with respect to the number of sub-reservoirs.  

\subsubsection{Shared Cross-Talk Matrix and the Reservoir Ring}

%A basic implementation of a parallel reservoir structure does not allow for cross-talk between reservoirs.  This is important because one sub-reservoirs' activity may benefit from information coming from other sub-reservoirs; for example spatially-located objects in a visual field, where sub-reservoirs take in information from 2D spatial location kernels.  If we wanted to fully connect each sub-reservoir, we could additionally add cross-connection matrices between sub-reservoirs. However, by adding  these recurrent weight matrices, we  lose the memory efficiency that we gained from splitting into sub-reservoirs.  Again dipping into deep learning, we can take inspiration from Convolutional Neural Networks \cite{lecun1998gradient}, which use shared convolutional weights. Here, we use a \emph{shared} recurrent weight matrix that all sub-reservoirs use to communicate with one another.  

In order to allow each sub-reservoir to exchange information, we introduce a novel meta-ring structure (M-Ring), wherein each sub-reservoir uses a shared weight matrix to connect to its left and right neighbors (Fig 1). Sub-reservoirs at  the end wrap around, creating a ring structure, which resembles the internal ring structure of simple RC networks \cite{rodan2010minimum}.   %A ring is chosen as the minimum complexity connectivity to connect all of the units, inspired by the ring structure of minimum complexity reservoirs \cite{rodan2010minimum}.  Of course, one could connect all sub reservoirs to all others, but this would increase the complexity of the network. 

\subsection{Deep Network Readout}
A centerpiece of our architecture is that unlike "traditional" RCs, the readout is done using a deep network using backpropagation \cite{rumelhart1995backpropagation}, which does credit assignment through the layers of the network via the chain rule.  While this does increase the computational complexity versus ridge regression, we argue that it is minimal; the backpropagation still stops before the recurrent layer, and does not need to be propagated backwards through time. The multiple parallel reservoirs are concatenated at the end, and fed in as a single layer to the input layer of the deep network. %While this is still done in the rate encoding framework with echo state networks, a purely event-driven schema is still possible with liquid state machines, plus backpropagation using spiking networks, of which many versions exist \cite{zenke2018superspike, lee2016training}.  This conversion to a purely event-based, spiking architecture is left for future work.

\subsection{Related Work}
There have been attempts to replace  the standard ridge regression in  echo state networks with alternatives [\cite{polydoros2015advantages}].  There also have  been efforts to apply CNN techniques to the reservoir readout [\cite{ma2019convolutional}].  Recent work has attempted to create hybrid models that integrate feed-forward networks into the reservoir structure \cite{zhao2020combining}. 

Other earlier works have attempted to attach a multi-layered backpropagation learning structure to an ESNs \cite{woodward2011reservoir, bianchi2017bidirectional}, however neither of these works used batch normalization or a parallel scheme.  One of the roadblocks hampering the use of reservoirs for real-world tasks is that the performance of the reservoir diminishes when the input dimension begins to increase: thus most test datasets used for reservoir computing, even very recently, involve very low-dimensional data.  Recently, it was shown in \cite{pathak2018model,qiao2016growing} that this issue can be alleviated by introducing a parallel reservoir schema, where the work is divided up among multiple reservoirs in parallel.  Our work extends this architecture for use with non-linear readouts and real-world data.  The idea of using reservoirs in parallel can be seen as an extension and fusion of the ideas of deep feed-forward multilayer perceptrons, where each neuron is replaced by a reservoir.  ESNs have also very recently been applied to EEG learning tasks \cite{jeong2020ear}.

\section{Experiments}
\label{S:2}

For consistency and reproducibility,  we provide  for all experiments, for each network type the results as the standard deviation and mean from multiple runs.  For all experiments, we test against the current state-of the art recurrent neural network gradient descent networks.  The two gradient-descent RNNs that we focus on are LSTMs \cite{hochreiter1997long} and GRUs \cite{cho2014learning}.  

%GRUs in particular are a more recent development
%, with work testing them on new datasets 
%\cite{wang2020prediction}. GRU cells are similar to LSTM cells except they lack a forget gate, which simplifies the architecture. However, one cannot state that GRUs are more advanced or better than LSTMs outright as LSTMs have advantages in certain scenarios \cite{weiss2018practical}, so we include both for comparison. 

To further stress test the comparisons, we also compare against deep (stacked) versions of LSTMs and GRUs with three recurrent stacked layers. We performed an exhaustive hyper-parameter search for the LSTM and GRUs to find the best parameters in each  experiment.

As ablation studies for the ring and parallel architectures, we include three variants of the HBP-ESN: 1) HBP-ESN, the parallel architecture without ring interconnections, 2) HBP-ESN M-Ring, the same but with the meta-ring interconnections as shown in Fig. 1, and 3) HBP-ESN Single, a non-parallel variant with one large reservoir.  We also include for both experiments a fully connected deep ANN (FCNN) as a baseline to demonstrate the necessity for a recurrent network for these tasks.  We split all data 80/20 for train/test. 

\begin{figure}
\begin{tikzpicture}
  \node (img1) {\includegraphics[width=\textwidth,height=\textheight,keepaspectratio]{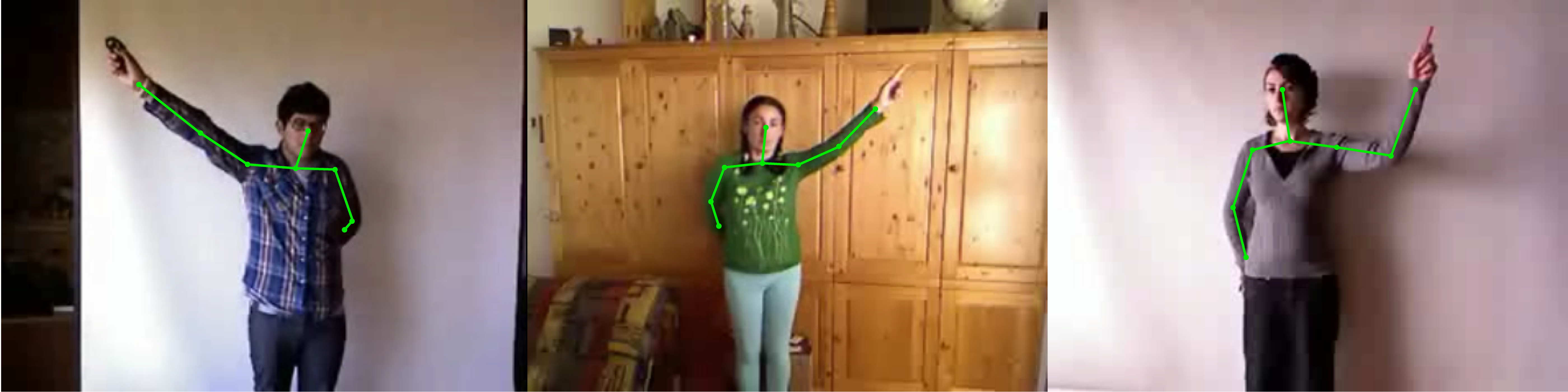}};
\end{tikzpicture}
\caption{ChaLearn Dataset: Helicoptor Airmashal Subset with 8 upper body skeleton  keypoints acquired from OpenPose containing 1792 samples - shown is label 3 (Take off) in Figure \ref{fig:helicopterairmarshal}. These examples demonstrate  variations  due to camera angles, body types, individual styles, hands used, etc. that make the gestures difficult to learn.}
\label{fig:openposekeypoints}

% \begin{tikzpicture}
%   \node (img2) {\includegraphics[width=\textwidth,height=\textheight,keepaspectratio]{gestures-small.png}};
% \end{tikzpicture}
% \caption{Keypoints index.}

\begin{tikzpicture}
  \node (img2) {\includegraphics[width=\textwidth,height=\textheight,keepaspectratio]{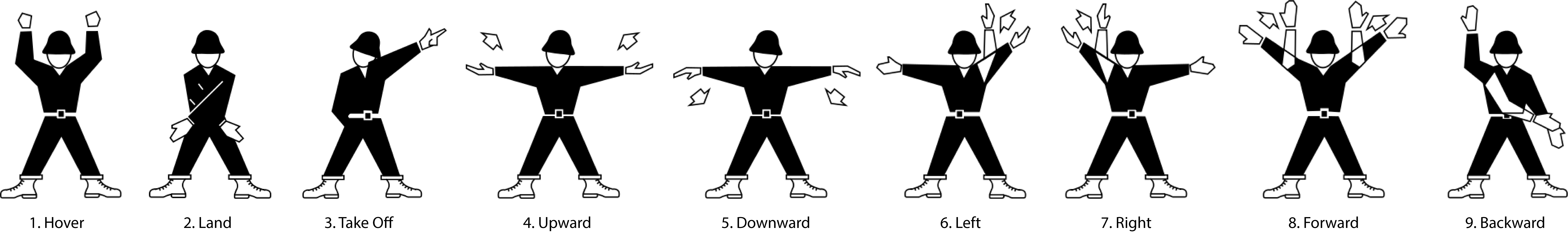}};
\end{tikzpicture}
\caption{Helicoptor Airmashal labels, a total of 9 classes, present in the ChaLearn dataset}
\label{fig:helicopterairmarshal}
\end{figure}

For our first experiment, we run a label classification on the ChaLearn \cite{wan2016chalearn} dataset, which extracts 8 upper body skeleton  keypoints using OpenPose \cite{cao2019openpose} for the  Helicoptor Aircraft Marshaling \cite{wiki2019aircraftmarshalling}. We train our networks to learn  classify these gestures into the 9 labels as shown in Figure \ref{fig:helicopterairmarshal}. 

For both experiments, we found  that batch norm performed better on the HBP-ESN in the fully-connected classification layers on these two datasets than for the LSTM or GRU.  Batch normalization was applied on the LSTM and GRU recurrent layers for those networks, but not the fully-connected readout layers, and batch norm was applied in the fully-connected readout layers of the HBP-ESNs, as these were the best performing network configurations for all networks.

%One conjecture might be that because the HBP-ESN reduces the deep learning problem to feed-forward learning of the reservoir states, and there is no learning in the recurrent layer, batch norm may be more effective, however further testing would be required to confirm this across more datasets. 

We can see from Table 1 that all variants of the HBP-ESN outperform the LSTM and GRU networks.  The sub-reservoir network was able to achieve comparable and even superior accuracy to the full reservoir, even though it uses an order of magnitude smaller weight matrices (Table 1).  We see that the M-Ring outperforms the vanilla parallel architecture as well. We found that for a problem as small as 16-dimensions, the backpropagation deep readout was able to make up for any shortcomings in separability issues as described in \cite{pathak2018model}.  However, the size of the reservoir made the training time much longer, and increasing the size of the reservoir further was not feasible in memory with one reservoir.  The parallel reservoir structure, however, performed as well as the single-reservoir with backpropagation.   All parallel versions used 8 sub-reservoirs.  The best HBP-ESN results came from sub-reservoirs of size 300 units, and from the M-Ring, size 400 units. For all gesture HBP-ESN experiments, for the reservoir a spectral radius was set to .1 according to the method of \cite{jaeger2001echo}, $\alpha$ was 0.05. For the backpropagation readout, a learning rate of 0.001 was used, weight decay was 0.001, and optimized with SGD with momentum 0.01 and ReLU activations. Batch size 64 was found to be the optimal, given the small dataset size.  For the reservoir, because it keeps a memory of past activity, only every 5th input frame was used to save on computation costs. $\beta$ was set to 0.005 for both experiments. 

\renewcommand{\arraystretch}{1.2}
\begin{table}[h!]
\begin{center}
\begin{tabular}{ |c|c|}
\hline
\textbf{Classification Methods} & \textbf{acc (\%)} $\pm$ \textbf{sd} \\
\hline
FCNN Baseline & 70.714 $\pm$ 2.519 \\
\hline
LSTM 1-Layer & 74.551 $\pm$ 2.542 \\
\hline
LSTM Deep & 73.770 $\pm$ 3.293 \\
\hline
GRU 1-Layer & 73.902 $\pm$ 3.282 \\
\hline
GRU Deep & 78.415 $\pm$ 2.364 \\
\hline
HBP-ESN Single &  \textbf{84.007} $\pm$ \textbf{0.853} \\
\hline
HBP-ESN &  83.763 $\pm$ 0.994 \\
\hline
HBP-ESN M-Ring &  \textbf{84.059} $\pm$ \textbf{1.090} \\
\hline
\end{tabular}
\end{center}
\caption{Airmarshal Gesture Classification Results.  Deep LSTM and GRU networks used 3 stacked recurrent layers. FCNN is the deep feed-forward ANN baseline.  HBP-ESN is the parallel scheme without interconnections; HBP-ESN Single is one large reservoir, and HBP-ESN M-Ring is the parallel structure with ring-connected components as shown in Fig. 1. Each number is the mean over 20 runs, with std. deviation posted as $\pm$.}
\end{table}

LSTM and GRU networks for 1-layer and deep versions had the best results for 256 units in the RNN layer(s) and 256 in the fully-connected 1-layer, and half of 256 in the subsequent layers for the deep version. The dropout was 0.5 in the input and 0.1 in the hidden layers both for RNN and fully-connected layers. As activation tanh was used for the RNN and ReLU for the fully-connected, no batch normalization, a learning rate of 0.001, the Adam optimizer with beta values of 0.9 and 0.999 and epsilon of 1e-07, and a batch size of 64 were used, which ran 20 times for all explorations.

%There appears to be limits to how small you can downsize the reservoir before you start to lose performance.  For example, for a desired neuron population of 3200, making 16 reservoirs of size 200 starts to see decreased performance versus 8 reservoirs of size 400.  There is a minimum reservoir size required for a sub-reservoir to capture even a one-dimensional time series.  This is consistent with the findings that even small input vectors require reservoirs of 100 nodes or more \cite{jaeger2005reservoir}. Thus, while we can compare the number of \emph{nodes} or \emph{neurons}, the number of \emph{synapses} for the parallel structure is far less.  

\begin{figure}[h!]
\begin{center}
\begin{tikzpicture}
  %%\node (img) {\includegraphics[scale=.5,keepaspectratio]{arousal.png}};
  \node (img) {\includegraphics[width=\textwidth,height=\textheight,keepaspectratio]{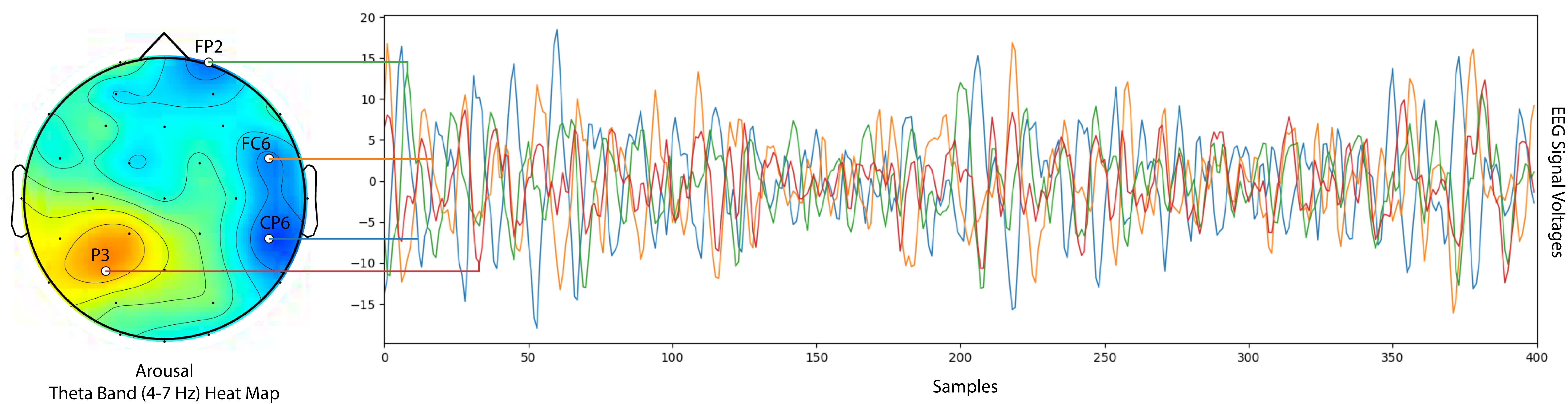}};
  %%\node[below=of img, yshift=1cm] {Recorded Time Steps};
  %%\node[left=of img, rotate=90, anchor=center, yshift=-1cm] {Recorded Signals};
\end{tikzpicture}
\caption{Left: A heatmap depicting the EEG nodes overlaid with the correlations in activity that correspond to the 'arousal' value.  Red values indicate positive correlation and blue negative.  For our purposes, we use all channels and let the network discern which channels are of importance.\cite{koelstra2011deap} Right: The raw EEG signal used for training, with lines showing the corresponding EEG node locations for example channels.}
\label{fig:deapEEGarousal}
\end{center}
\end{figure}

For our second experiment, we run a regression learning on the DEAP EEG dataset \cite{koelstra2011deap} (Fig. \ref{fig:deapEEGarousal}).  We use all 40 channels for training corresponding to the raw sensor recordings from 32 EEG electrodes and 8 body physiological sensors.  The dataset has recordings from 32 participants, each with 40 trials, watching 1-minute video clips, where EEG and bio-sensor recordings were taken during the video watching session.  Afterwards, participants were asked to rate emotions (valence, arousal, dominance, liking) on a 9-point scale. As in \cite{pan2020eeg}, we break up each 60-second trial into 1-second epochs, corresponding to 134-length vectors; each 1 second epoch is given the same label as the corresponding 60 second trial, giving 2400 samples per participant.  Unlike most papers that run classification on high/low values, we run pure regression on the numerical label value and report results  after running the network 5 times and showing the mean and std. dev. in Table 2.  Lastly, shown here are the regression results for "arousal", corresponding to the second index of the label array. Other labels and combined regression results will be shown in supplementary materials. All input data was channel-wise normalized between -1 and 1. Thus at each of 134 time steps, a 40-element vector was fed into the networks corresponding to the 40 channels, and the network output was regressed to the arousal label. For all HBP-ESN runs for Experiment 2, 140 total reservoir neurons were used: 140 for the single reservoir, and 4 sub-reservoirs of size 40 were used for the parallel schemes. As Table 2 shows, the M-Ring variant outperformed all other networks in terms of both accuracy and std. deviation. 

\renewcommand{\arraystretch}{1.2}
\renewcommand\bottomfraction{.9}
\begin{table}[h!]
\begin{center}
\begin{tabular}{ |c|c|  }
\hline
%\multirow{2}{*}{\textbf{Method}} & \multicolumn{2}{c|}{\textbf{Within}} & \multicolumn{2}{c|}{\textbf{Between}} \\
\textbf{Regression Methods} & \textbf{mse} $\pm$ \textbf{sd} \\
\hline
FCNN & 2.402 $\pm$ 0.164 \\
\hline
LSTM Deep & 1.794 $\pm$ 0.307 \\
\hline
GRU Deep & 3.100 $\pm$ 0.792 \\
\hline
HBP-ESN Single & \textbf{1.320} $\pm$ \textbf{0.105} \\
\hline
HBP-ESN & 1.457 $\pm$ 0.119 \\
\hline
HBP-ESN M-Ring & \textbf{1.433} $\pm$ \textbf{0.122} \\
\hline
\end{tabular}
\end{center}
\caption{Results from the EEG emotion regression learning task for the 'arousal' label.  Each network was ran individually on all 32 participants 5 separate times, and mean and std. deviation across all runs are posted.}
\vspace{-4mm}
\end{table}

At each layer of the deep learning readout, Batch Normalization \cite{ioffe2015batch} is applied, which regularizes each mini-batch.  %This technique has shown great success in the deep learning literature, and being able to use it is one of the benefits of using a deep learning readout mechanism. 
The batch size we use for our experiments is 64, found to be the best from empirical testing.  
%In addition, use of deep learning gives us access to its years of work into normalization techniques.  
Here we apply weight decay \cite{smith2018disciplined} as our main normalization technique, as it can empirically outperform dropout \cite{dahl2013improving}, another common deep learning normalization technique.  To save on memory usage, only every other reservoir state is fed into the reservoir, so the number of reservoir states is downsampled by two. For the LSTM and GRU network comparisons, dropout is used instead, %after testing with batch normalization and dropout, and determining that dropout 
which was found   more effective for these networks. 

%\section{Discussion and Limitation}
%One of the main limitations of ESNs 

%that is not addressed in this work, but will be addresed in future work,

%is that the data fed into the readout is the entire time history of the reservoir.  This is the "standard" way to learn with ESNs \cite{jaeger2001echo}, but less biologically plausible.
%as this would require the brain to keep a time history of all past inputs. 
%However, even among standard regression-learning ESNs, online learning techniques exist \cite{lukovsevivcius2012practical}.  In terms of the HDN-ESN, online learning is even more feasible, as batch learning with stochastic gradient descent is already set up to allow for incremental learning.  In future work we will study mechanisms to allow for online incremental batch learning of the reservoir states. 

%One further area of improvement is to reduce the variance of the HBP-ESN.  We believe this arises mainly out of the random initial matrix, whose behavior depends much more on initial conditions than gradient-learning RNNs, as the HBP-ESN relies on the chosen random initial matrix for all future learning.  While we did not perform reservoir weight optimizations for this work for sake of clarity, future work will incorporate these optimizations to reduce the variance. 

\section{Conclusion}

We demonstrated the use of a backpropagation hybrid mechanism for parallel reservoir computing with a meta ring structure and its application on a real-world gesture recognition dataset.  We show that it can be used as an alternative to state of the art recurrent neural networks, LSTMs and GRUs.  Future work will expand upon and further integrate deep learning techniques into the reservoir learning framework.  We offer this novel network as a new route to learning temporally changing or sequential data in a way that utilizes random recurrent matrices without the use of Backpropagation Through Time.  We believe this can form the building blocks for future architectures that can modularize the reservoirs for hierarchical systems.

\begin{ack}

This work was supported by the University of Maryland COMBINE program NSF award DGE-1632976 and NSF grant BCS 1824198. The authors would like to also thank Vaishnavi Patil at the University of Maryland for thoughtful conversation and advice.

\end{ack}

\section{Broader Impacts}

We believe that bio- and brain-inspired neural networks have the  potential to both improve the state of the art of A.I., but also to give insights into how the brain  operates. While much of the brain remains a mystery, even potentially non-biological advances in AI, which are inspired by neuroscience can have big impacts  on how researchers orient their theories of the brain.  For example, while we are not arguing for or against the plausibility of backpropagation, the idea of error reduction from backpropagation networks has led to new theories about the central role of error reduction in the brain \cite{marblestone2016toward}.  Our hope is that by proposing  biologically inspired networks, such as our hybrid network here that avoids mechanisms that are generally agreed upon as biologically implausible like BPTT \cite{manneschi2020alternative}, we can add further evidence for ways the brain can process temporal information.  In addition, by testing on EEG datasets, we also aim to show  that our work can be of broader use to the medical community in identifying brain patterns for a wide variety of uses. One example of a future direction we hope to take this work is for  anomaly detection from brain signals to help identify early signs of epileptic seizures.

%\section*{References}

%\bibliographystyle{plain}
\bibliography{biblio}
%{\small
%\bibliographystyle{abbrvnat}
%\bibliography{biblio.bib}
%}

\end{document}